\documentclass{bmvc2k}
\usepackage{color,soul}
\usepackage{multirow}
\usepackage{multicol}
\usepackage{amssymb} 
\usepackage{amsmath}

\title{CoMoGCN: Coherent Motion Aware Trajectory Prediction with Graph Representation}
\addauthor{Yuying Chen*}{ychenco@connect.ust.hk}{1}
\addauthor{Congcong Liu*}{cliubh@connect.ust.hk}{1}
\addauthor{Bertram Shi}{eebert@ust.hk}{1}
\addauthor{Ming Liu}{eelium@ust.hk}{1}

\addinstitution{
 Robotics Institute\\
 Hong Kong University of Science and Technology\\
 Hong Kong, China \\ 
}

\runninghead{Chen, Liu, Shi, Liu}{Coherent Motion Aware Trajectory Prediction WITH GCN}

\begin{document}

\maketitle
\begin{abstract}
Forecasting human trajectories is critical for tasks such as robot crowd navigation and autonomous driving. 
Modeling social interactions is of great importance for accurate group-wise motion prediction. 
However, most existing methods do not consider information about coherence within the crowd, but rather only pairwise interactions. 
In this work, we propose a novel framework, coherent motion aware graph convolutional network (CoMoGCN), for trajectory prediction in crowded scenes with group constraints. First, we cluster pedestrian trajectories into groups according to motion coherence. 
Then, we use graph convolutional networks to aggregate crowd information efficiently. 
The CoMoGCN also takes advantage of variational autoencoders to capture the multimodal nature of the human trajectories by modeling the distribution.
Our method achieves state-of-the-art performance on several different trajectory prediction benchmarks, and the best average performance among all benchmarks considered.
\end{abstract}

\section{Introduction}
\label{sec:intro}
Forecasting human trajectories is of great importance for tasks, such as robot navigation in crowds, autonomous driving, and crowd surveillance.
For autonomous robot systems, predicting the human motion enables feasible and efficient planning and control.

However, making accurate trajectory predictions is still a challenging task because pedestrian trajectories can be affected by many factors, such as the topology of the environment, intended goals, and \textit{social relationships and interactions} \cite{rudenko2019human}. Furthermore, the highly \textit{dynamic} and \textit{multimodal} properties inherent in human motion must also be considered.


Multimodality in trajectory prediction has been studied recently \cite{gupta2018social,sadeghian2019sophie,kosaraju2019social,lee2017desire,amirian2019social}. Most past work uses generative adversarial models (GANs) to generate multiple predictions. 
However, GANs suffer from the instability of adversarial training, which is sensitive to hyperparameters and structure \cite{wang2019generative}. 
As an alternative, variational autoencoder (VAE) is relatively more stable. Lee \textit{et al.} present a CVAE based framework to predict future object locations \cite{lee2017desire}. A recent work adopted CVAE for trajectory prediction \cite{ivanovic2019trajectron}. 
This paper takes advantage of the VAE to capture the multimodality of human trajectories.

Recently, some works have proposed to model the dynamic interactions of pedestrians by combining information from pairwise interactions, through pooling mechanisms such as max-pooling \cite{gupta2018social} and self-attention pooling \cite{sadeghian2019sophie}.
However, those works do not completely capture important information about the geometric configuration of the crowd. Furthermore, these works rely on some ad-hoc rules to handle varying numbers of agents, such as setting a maximum on the number of agents and using dummy values for non-existing agents \cite{sadeghian2019sophie}.
To avoid such ad-hoc assumptions, Chen \textit{et al.} \cite{chen2020robot} propose to use graph convolutional networks (GCN) to aggregate information about neighboring humans for robot crowd navigation tasks. 
The GCN can handle varying numbers of neighbors naturally, and can be extended to modulate interactions by changing its adjacency matrix. In this paper, we use a similar graph structure for crowd information aggregation in a different task: trajectory prediction.

Most previous work has focused only on the interactions between pairs of humans. 
Coherent motion patterns of pedestrian groups, which encode rich information about implicit social rules, has rarely been considered. 
This lack of attention may be due in part to the lack of information about social grouping in current benchmark datasets, such as the commonly used \textbf{ETH \cite{pellegrini2009you} and UCY \cite{lerner2007crowds}} datasets, for trajectory prediction. 
To address this unavailability, we add coherent motion cluster labels to trajectory prediction datasets using a coherent filtering method \cite{zhou2012coherent}, and leverage DBSCAN clustering to compensate for the drawbacks of the coherent filtering method in the small group detection. These coherent motion labels provide a mid-level representation of crowd dynamics, which is very useful for crowd analysis.
We incorporated the coherent motion constraints into our model by using GCNs for intergroup and intragroup relationship modeling.

There are several main contributions of our work:
\begin{itemize}

\item We introduce graph convolutional networks (GCN) to better model social interactions within human crowds. The use of GCNs enables our approach to handle varying crowd sizes in a principled way. Interactions between humans can be controlled easily by modifying the adjacency matrix.

\item Unlike past work that considered pairwise interactions between individuals only, we take into account coherent motion constraints inside crowds to better capture social interactions. 

\item We developed a hybrid labeling method to add coherent motion labels to trajectory prediction datasets. We will release the re-labelled dataset publicly for use by other researchers\footnote{https://sites.google.com/view/comogcn}.

\item We take advantage of the VAE to handle multimodality in trajectory modeling. 

\item With the above mechanisms, the CoMoGCN achieves state-of-the-art performance on several different trajectory prediction benchmarks, and the best average performance across all datasets considered.

\end{itemize}
\section{Related works}
\subsection{Crowd Interaction}
A pioneering work for crowd interaction modeling, the Social Force Model (SFM) proposed by \cite{helbing1995social}, has been applied successfully to many applications such as abnormal crowd behavior detection\cite{mehran2009abnormal} and multi-object tracking \cite{pellegrini2009you}.
However, as discussed in \cite{alahi2016social}, the social force model can model simple interactions, but fails to model complex crowd interactions. 
There are also other hand crafted feature based models, such as continuum dynamics \cite{treuille2006continuum}, discrete choice \cite{antonini2006discrete} and Gaussian Process models \cite{wang2007gaussian}.
However, all the above methods are based on hand-crafted energy functions and specific rules, which limit their performance. 


\subsection{RNN for Trajectory Prediction}
Recently, Recurrent Neural Networks (RNN), such as the Long Short Term Memory (LSTM), have achieved many successes in trajectory prediction tasks \cite{alahi2016social,su2017forecast,xu2018encoding,zhang2019sr,lisotto2019social,hasan2018mx}. 
Alahi \textit{et al.} proposed a social pooling layer to model neighboring humans \cite{alahi2016social}. 
Gupta \textit{et al.} proposed a pooling module, which consists of an MLP followed by max-pooling to aggregate information from all other humans \cite{gupta2018social}.
Sadeghian \textit{et al.} \cite{sadeghian2019sophie} adopted a soft attention module to aggregate information across agents.
More recent work uses GCNs to aggregate information by treating humans as nodes and modeling interaction through edge strength for robot navigation \cite{chen2020robot}. 
Similarly, a variant of the GCN, the Graph Attention Network (GAT), has been used to model the social interactions \cite{kosaraju2019social}. 
However, the use of multi-head attention in the GAT increases the number of parameters and the computational complexity of the GAT in comparison to the GCN. 
In this work, we integrate information across humans using GCNs, which enables our method to handle varying crowd sizes.


\subsection{Coherent Motion Information for Motion Prediction}

Most previous work only pay attention to interactions among pairs of pedestrians. However, the pedestrian trajectories are also influenced by more complex social relations between humans. 
Coherent motion patterns inside crowds, which encode implicit social information, have been shown to be useful in many applications, such as crowd activity recognition\cite{wang2008unsupervised}. 
Bisagno \textit{et al.} \cite{bisagno2018group} considered intragroup interactions for trajectory predictions, but neglected intergroup interactions.
Current benchmark datasets for trajectory prediction do not provide coherent motion labels.

Several works have been done in detecting coherent motions \cite{zhou2012coherent} and measuring the collectiveness of crowds \cite{mei2019measuring}. 
Zhou \textit{et al.} \cite{zhou2012coherent} proposed the coherent filtering that detects invariant neighbors of every individual, and measures the velocity correlations for motion clustering. It shows good performance on collective motion benchmark and can detect coherent motions given the crowd trajectories in a short time window. In this paper, we use the coherent filtering method to label trajectory prediction datasets. In addition, we leverage DBSCAN clustering to compensate for the disadvantages of the coherent filtering method in small group detection. Based on the labels, we incorporate the coherent motion information into our model for better interaction modeling.


\section{Method}

\begin{figure*}[htb]
  \centering
  \includegraphics[width=1.0\columnwidth]{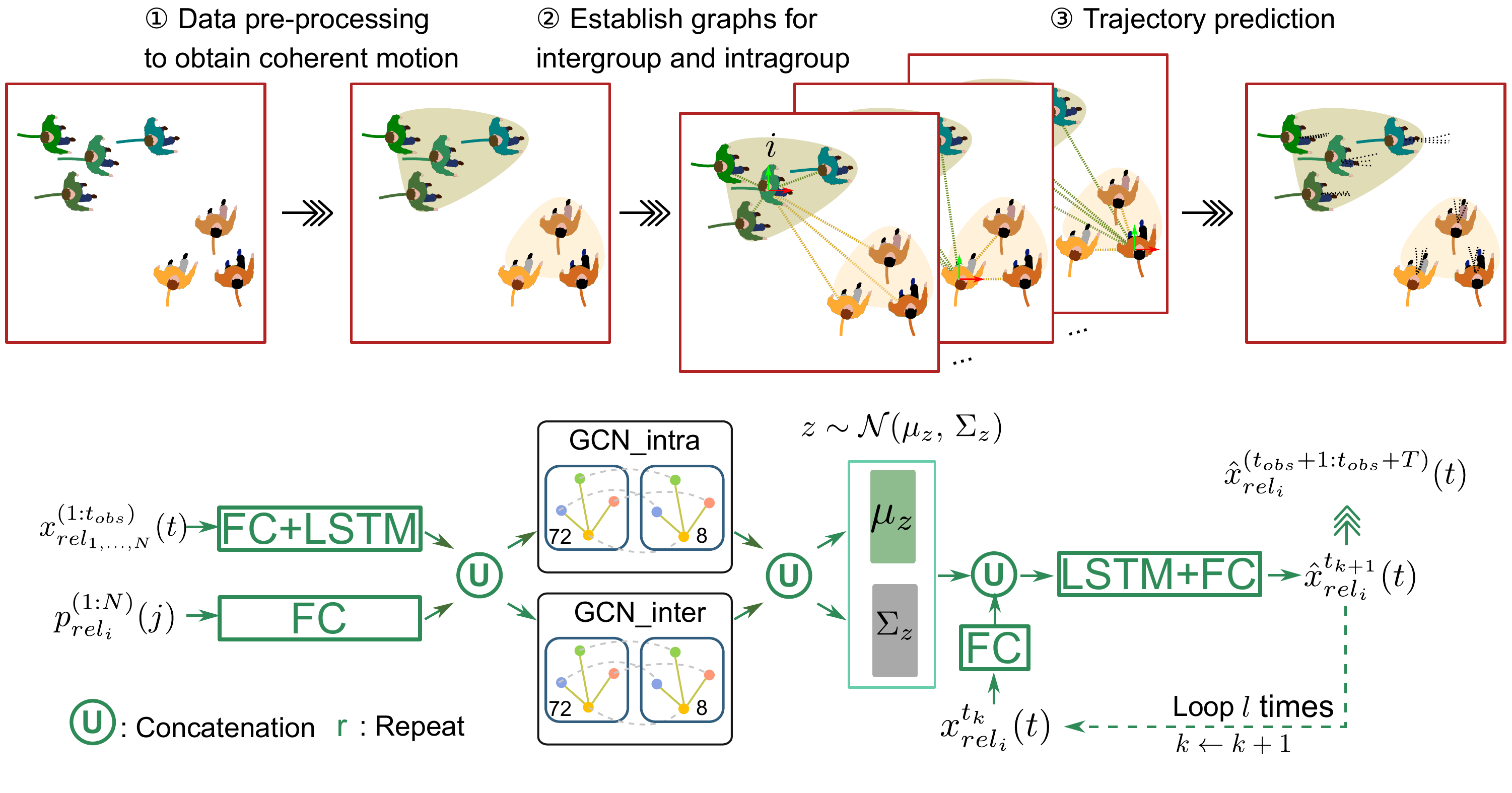}
  \caption{System overview. There are three procedures: 1. We obtain coherent motion labels for each human in an offline data pre-processing procedure. 2. Based on the coherent motion labels for each human, we establish graphs capturing intergroup and intragroup relationships. The encoder LSTM takes past trajectories as input and feeds the encoded features into two GCNs. 
  3. The embeddings from the two GCNs are concatenated and forwarded to an MLP to create a distribution with mean $\mu_z$and variance $\Sigma_z$. Then, features are sampled from the distribution and fed into a decoder LSTM for trajectory prediction.\label{fig:structure} }
\end{figure*}
 







\subsection{Problem Definition}
The goal of this work is to generate the future trajectories of all humans in a scene at the same time.
The trajectory of a person $i$ is defined using $x_{rel_{i}}^t = (x_i^t, y_i^t)$ which denotes the relative position of human $i$ at time step $t$ to the position at $t-1$.
Consistent with previous works \cite{gupta2018social,sadeghian2019sophie}, the observed trajectory of all humans in a scene is defined as $x_{rel_{1,...,N}}^{(1:t_{obs})}$ for time steps $t = 1,...t_{obs}$; the future trajectory to be predicted is defined as  $x_{rel_{1,...,N}}^{(t_{obs}+1:t_{obs}+T)}$ for time step $t = t_{obs} + 1,..., t_{obs} + T$, where the number of humans $N$ may change dynamically. The model aims to generate trajectories $\hat{x}_{rel_{1,...,N}}^{(t_{obs}+1:t_{obs}+T)}$ whose distribution matches that of ground truth future trajectories of all humans $x_{rel_{1,...,N}}^{(t_{obs}+1:t_{obs}+T)}$.

\subsection{Overall Model}
Figure \ref{fig:structure} shows the overall framework of our method for trajectory prediction. 
Data pre-processing is applied offline to obtain the coherent motion pattern for each human.
For feature extraction, we first use a single layer MLP (FC) to encode each pedestrian's relative displacements as a fixed-length embedding. These embeddings are fed to an LSTM as shown below: 
\begin{equation}
\label{eqn:encodelstm}
e_i=LSTM_{en}(MLP_{enc}(x_{rel_i};W_{enc}),h_{enc_i},W_{en})
\end{equation}
where $W_{enc}$ is the weight of FC layer, and $W_{en}$ is the weight of the encoding LSTM.
On the other hand, for specific person $i$, the relative position of other humans are fed into an FC layer to obtain social information $p_i$ which is similar to the pooling module in Social GAN \cite{gupta2018social}.

Then the features from all pedestrians $e_{1, ..., N}$ and the interested person's social information $p_i$ are concatenated together as the input to the two GCNs for intergroup and intragroup interaction aggregation:
\begin{align}
\label{eqn:GCN}
V_{\text{intra}_i} =& GCN_{\text{intra}}([e_{1, ..., N},p_i], A_{\text{intra}}, W_{\text{intra}})\\
V_{\text{inter}_i} =& GCN_{\text{inter}}([e_{1, ..., N},p_i],A_{\text{inter}}, W_{\text{inter}})
\end{align}
where $A_{\text{intra}}$ and $A_{\text{inter}}$ denote the adjacency matrices as described in more detail in Section. \ref{sec:coherent_incor}.
$W_{\text{intra}}$ and $W_{\text{inter}}$ are weight matrices.
We extract the feature of node $i$ after the final graph convolutional layer as the features $V_{intra_{i}}$ and $V_{inter_{i}}$.  

The features computed by the outputs of the two GCNs are then concatenated together and input to an MLP, which computes the mean and variance of a distribution over the feature vectors to be input to the decoder: 
\begin{equation}
\label{eqn:vae}
\mu_z, \Sigma_z=MLP_{\text{vae}}([V_{\text{intra}_i}, V_{\text{inter}_i}], W_{\text{vae}})
\end{equation}
\noindent where $W_{\text{vae}}$ is the weight matrix. 
We sample an input feature vector to the decoder stage, $z$, from this distribution $z\sim\mathbf{N}(\mu_z,\Sigma_z)$ and concatenate it with the embedding computed from an embedding of the last predicted state. The resulting features $c$ are fed into the decoder LSTM cell for trajectory prediction:
\begin{equation}
\label{eqn:decodelstm}
\hat{x}_{rel_i}=MLP_{dec}(LSTM_{de}(c, h_{de_i};W_{de});W_{dec})
\end{equation}
\noindent where $W_{de}$ is the weight for decoder LSTM and $W_{dec}$ is the weight for decoder MLP.

\subsection{Coherent Motion Clustering for Pedestrian Groups}
For coherent motion detection, we use the coherent filtering proposed by \cite{zhou2012coherent}. The process takes the positions of humans from consecutive frames $t_1$ to $t_k$ and generates a clustering index for each human and for each frame. Humans sharing the same index are considered to have coherent motion. The process of coherent filtering mainly includes three steps: a) finding K nearest neighbors b) finding the invariant neighbors of a individual c) measuring the time-averaged velocity correlations of the invariant neighbors to the individual. Among these individual-neighbor pairs, pairs with correlation intensity above a threshold are marked as coherent pairs.

Though this method is effective for crowds with large crowd densities, it performs poorly for sparse crowds and fails to detect small groups. To compensate, we apply an extra clustering step, the DBSCAN method \cite{ester1996density}, for the unlabeled humans. As a density based clustering method, it relies on the distance to find the neighbors. We account for moving direction and calculate the angular distance of each pair of humans. These differences are used to classify humans into clusters. 

Our hybrid labeling method improves the labeling yield and generates better labels than the coherent filtering alone.  
Figure 1 of the supplementary file shows examples of detection by coherent filtering on each dataset. 
The quantitative evaluations of the coherent filtering and of our hybrid labeling method are shown in Table 2 and 3 of the supplementary file.
Figure 2 of the supplementary file shows a qualitative comparison between the coherent filtering and our method. The parameter settings are shown in Table 1 of the supplementary file. 

\subsection{Graph Convolutional Networks}

Dealing with the large and varied numbers of humans in a scene is one of the main challenges for multi-human trajectory prediction. Previous works adopted ad-hoc solutions such as setting a maximum number of humans\cite{sadeghian2019sophie}. 
In this work, we address this problem in a simpler and more principled way through graph representations. 
Nodes in the graph denote humans in the crowd. 
In the following, we denote the number of humans in the crowd by $N$.

We adopt a two-layer graph convolutional networks (GCNs) \cite{kipf2016semi} to aggregate information in crowds. 
To each node in the network, we associate a feature vector, which contains important information about the node. 
The graph convolutional layer is the main building block of GCNs. 
It takes input feature vectors for each node and converts them to output feature vectors for each node by integrating information both within and across nodes. 
We use $I$ to denote the dimension of the input feature vectors and $O$ to denote the dimension of the output feature vectors
The input feature vectors of layer $l$ are represented by matrix $\mathbf{H}^l \in \mathbb{R}^{N \times I}$. 
The input feature matrix is converted to output vectors represented by a matrix $\mathbf{H}^{l+1} \in \mathbb{R}^{N \times O}$ based on the layer-wise forward rule:
\begin{equation}
\label{eqn:update}
\mathbf{H}^{l+1} = \sigma \left ( \boldsymbol{A} \mathbf{H}^l \mathbf{W}^l \right )
\end{equation}
$\mathbf{W}^l \in \mathbb{R}^{I\times O}$ is a trainable weight matrix for layer $l$.  $\boldsymbol{A} \in \mathbb{R}^{N \times N}$ is the adjacency matrix of the graph, whose values determine how information from different nodes is aggregated.
Each row of $\boldsymbol{A}$ is normalized to sum to one. 
$\sigma(\cdot)$ is Relu activation function. 

\begin{figure}[htb]
  \centering
  \includegraphics[width=0.75\columnwidth]{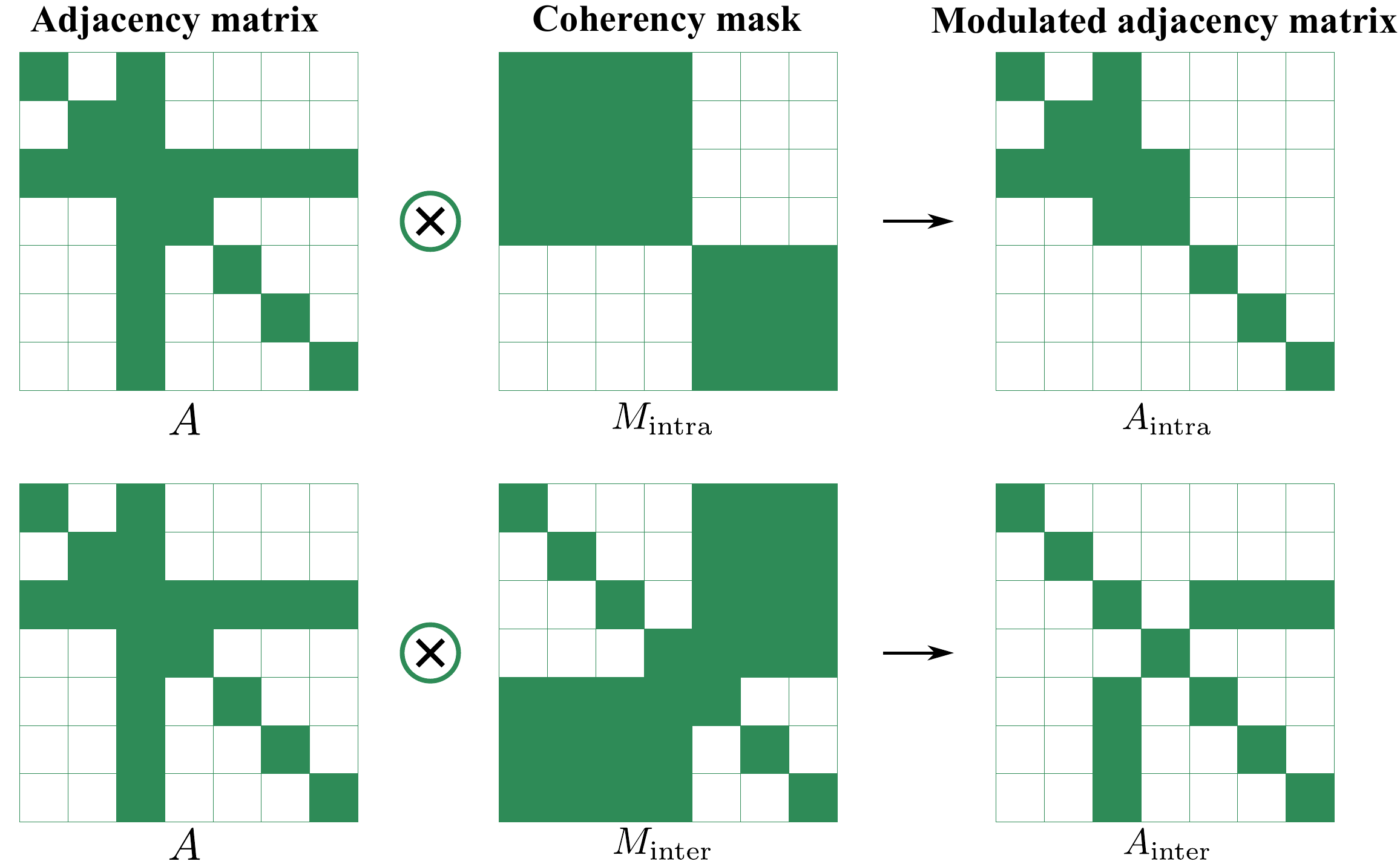}
  \caption{An example of how the adjacency matrices of the GCNs for crowd information aggregation and determined. The example considers the adjacency matrix for the GCNs of human i = 3, who is in the same cluster as humans 1,2 and 4, but not humans 5, 6 and 7. \label{fig:adj} }
\end{figure}

The adjacency matrix reflects the connections between nodes of the graph.
The vanilla GCN assumes that the qualitative influence of each human on another (as determined by $\mathbf{W}^l$) is the same and only the strength of that influence can be modulated (through the adjacency matrix). However, we think that the qualitative effect of humans in the crowd on a particular human's trajectory are different, based on whether the humans are in the same group or not. A single GCN can not handle this. Thus, we propose to use two GCNs.
As shown in Fig.\ref{fig:adj}, for each human, we modulate the adjacency matrix by multiplication with two coherence masks which encodes the intergroup and intragroup labels.
Then we obtain two adjacency matrix denoting intergroup connection ($A_{\text{inter}}$) and intragroup ($A_{\text{intra}}$) connection separately for each human by pixelwise multiplying the adjacency matrix ($A$) with the masks.
We set the value in the adjacency matrix by first constructing a binary matrix denoting connections between nodes, and then normalizing each row.


By modulating the adjacency matrix of GCNs with coherent motion information, we incorporate implicit social relations into our network for better interaction modeling.
\label{sec:coherent_incor}

\subsection{Implementation Details}
We trained the network with Adam optimizer. The mini-batch size is 64 and the learning rate is 1e-4. The models were trained for 200 epochs.
The encoder encodes the relative trajectories by a single layer of MLP ($MLP_{enc}$) with dimension of 16 followed by an LSTM ($LSTM_{en}$)with a hidden dimension of 32. The embedding output from LSTM was then concatenated with the features extracted from relative position from other humans by a single MLP with dimension of 16.
The concatenated features are then fed into two GCNs for feature integration.
Then hidden number for two graph convolutional layer has the dimension of 72 and 8 separately.
Then an MLP ($MLP_{vae}$) was used to take state of humans to create a distribution with mean and variance. Then we sample $z$ from this distribution with a dimension of 8, and fed it into an LSTM ($LSTM_{de}$)with dimension of 32 and followed by an MLP ($MLP_{dec }$)with dimension of 2 for decoding.

\section{Experiments}
In this section, we evaluate our method in two public datasets ETH \cite{pellegrini2009you} and UCY \cite{lerner2007crowds}.
The ETH datasets contain two scenes (ETH and Hotel) while the UCY datasets contain three scenes (Zara1, Zara2, and Univ). 
There are five sets of data with four different scenarios and 1536 pedestrians in total.

\subsection{Evaluation Methodology}
Following the setting in \cite{gupta2018social}, we adopt the leave-one-out approach, i.e. train with four sets and test in the remaining set.
We take trajectories with 8 time steps as observation and evaluate trajectory predictions over the next 12 time steps.

\subsubsection{Metrics}
Similar to previous works \cite{gupta2018social,sadeghian2019sophie,kosaraju2019social}, we adopt two standard metrics including Average Displacement Error (ADE) and Final Displacement Error (FDE) in meter. 

\textit{ADE}: Mean L2 distance between ground truth and predictions of all time steps.

\textit{FDE}: Mean L2 distance between ground truth and prediction at the final time step.

\subsubsection{Baselines}
We compare our work with following several recent works based on generative models:


\textit{Social GAN (S-GAN)} \cite{gupta2018social}: A generative model using GAN to generate multimodal predictions. It utilizes a global pooling module to combine crowd interactions by an MLP followed by a max-pooling layer.

\textit{Sophie} \cite{sadeghian2019sophie}: A improved GAN based model which considers both social interactions and physical interaction with scene context.

\textit{Trajectron} \cite{ivanovic2019trajectron}: A generative model based on CVAE for multimodal predictions with spatiotemporal graphs. 

\textit{Social-BiGAT} \cite{kosaraju2019social}: A generative model using Bicycle-GAN for multimodal prediction and GAT for crowd interaction modeling.

\subsection{Quantitative results}

\begin{table*}[]
\resizebox{\columnwidth}{!}{

\begin{tabular}{|l|cccc|ccccc|}
\hline
        & \multicolumn{4}{c|}{Baselines}                                                                                  & \multicolumn{5}{c|}{Ours}                                                                                                                                                                                                        \\ \cline{2-10} 
Dataset & \multicolumn{1}{c|}{S-GAN} & \multicolumn{1}{c|}{Sophie} & \multicolumn{1}{c|}{Trajectron} & Social-BiGAT       & \multicolumn{1}{c|}{MLP} & \multicolumn{1}{c|}{GCN} & \multicolumn{1}{c|}{GAT} & \multicolumn{1}{c|}{\begin{tabular}[c]{@{}c@{}}GCN+group\\ (CF)\end{tabular}} & \begin{tabular}[c]{@{}c@{}}GCN+group\\ (Hybrid)\end{tabular} \\ \hline
ETH     & 0.81/1.52                  & 0.70/1.43                   & \textbf{0.59/1.17}              & 0.69/1.29          & 0.73/1.40                & 0.72/1.31                & 0.73/1.36                & 0.71/1.28                                                                     & 0.70/1.26                                                       \\
HOTEL   & 0.72/1.61                  & 0.76/1.67                   & 0.42/0.80                       & 0.49/1.01          & 0.45/0.93                & 0.41/0.81                & 0.41/0.85                & 0.37/0.76                                                                     & \textbf{0.37/0.75}                                              \\
UNIV    & 0.60/1.26                  & 0.54/1.24                   & 0.59/1.21                       & 0.55/1.32          & 0.61/1.31                & 0.55/1.18                & 0.55/1.19                & 0.55/1.19                                                                     & \textbf{0.53/1.16}                                              \\
ZARA1   & 0.34/0.69                  & 0.30/0.63                   & 0.55/1.09                       & \textbf{0.30/0.62} & 0.34/0.72                & 0.35/0.74                & 0.35/0.74                & 0.34/0.72                                                                     & 0.34/0.71                                                       \\
ZARA2   & 0.42/0.84                  & 0.38/0.78                   & 0.52/1.04                       & 0.36/0.75          & 0.33/0.71                & 0.32/0.68                & 0.31/0.68                & 0.32/0.68                                                                     & \textbf{0.31/0.67}                                              \\ \hline
AVG     & 0.58/1.18                  & 0.54/1.15                   & 0.53/1.06                       & 0.48/1.00          & 0.49/1.01                & 0.47/0.94                & 0.47/0.96                & 0.46/0.93                                                                     & \textbf{0.45/0.91}                                              \\ \hline
\end{tabular}
}
\caption{Quantitative results. We adopted two metrics Average Displacement Error (ADE) and Final Displacement Error (FED) for evaluation over five different datasets (ADE/FDE in meters). Our full model (GCN +group (hybrid)) achieves state-of-the-art results outperforming all baseline methods (lower value denotes better performance).}
\label{tab:pred_error}
\end{table*}

\subsubsection{Comparison to state-of-the-art methods}
As shown in Table \ref{tab:pred_error}, we compare our models with various baselines. The average displacement error (ADE) and final displacement error (FDE) were reported across five datasets.
Following settings in every baseline, we run 20 samples for evaluation.

It is clear to see that our final model with GCN and coherent motion constraints beat all baselines and obtain more consistent results in both ADE and FDE. 
Compared to Social GAN, we achieve 22.4\% improvement in ADE and 22.9\% improvement in FDE on average.
Compared to Sophie who use additional scene context information, we achieve 16.7\% improvement in ADE and 20.9\% improvement in FDE on average.
Compared to Trajectron who also uses VAE as backbone network, we achieve 15.1\% improvement in ADE and 14.2\% improvement in FDE on average.
Compare to Social-BiGAT who also considers graph structure for interaction modeling, we achieve 6.3\% improvement in ADE and 9.0\% improvement in FDE on average.

\subsubsection{Ablation study}
We conduct several ablation studies to validate the benefits of the use of GCN and coherent motion information. 

To show the benefit of the use of GCN, we investigated another model that replaces GCN with MLP (followed by max-pooling, similar to the pooling module in social GAN \cite{gupta2018social}) as shown in Table \ref{tab:pred_error}.

When comparing the model using GCN with MLP, we can see that the one with GCN achieves 4.1\% improvement in ADE and 6.9\% improvement in FDE.

To show the benefit of the incorporation of coherent motion information, we compare our full model with the one without considering coherent information (only using GCN), and GAT (same implementation with \cite{velivckovic2017graph}).

When compare the full model with the one using GCN only, we can see that our full model with coherent motion information achieves 4.3\% improvement in ADE and 3.2\% improvement in FDE. 
When compare the full model with GAT, we can see that the full model achieves 4.3\% improvement in ADE and 5.2\% improvement in FDE.

The above ablation studies clearly demonstrate the benefits of the use of GCN and the introduction of coherent motion information.

We further investigated trajectory prediction performance of models with different coherent detection method, Coherent Filtering method (CF) \cite{zhou2012coherent} vs. our hybrid labeling method (hybrid). We can see that model with our hybrid coherent detection method (Coherent Filtering + DBSCAN) outperforms model with Coherent Filtering method by 2.2 \% improvement in ADE and 2.2 \% improvement in FDE on average. The improvements are consistent over all five datasets.
\subsection{Qualitative results}
\begin{figure*}
  \centering
  \includegraphics[width=1.0\columnwidth]{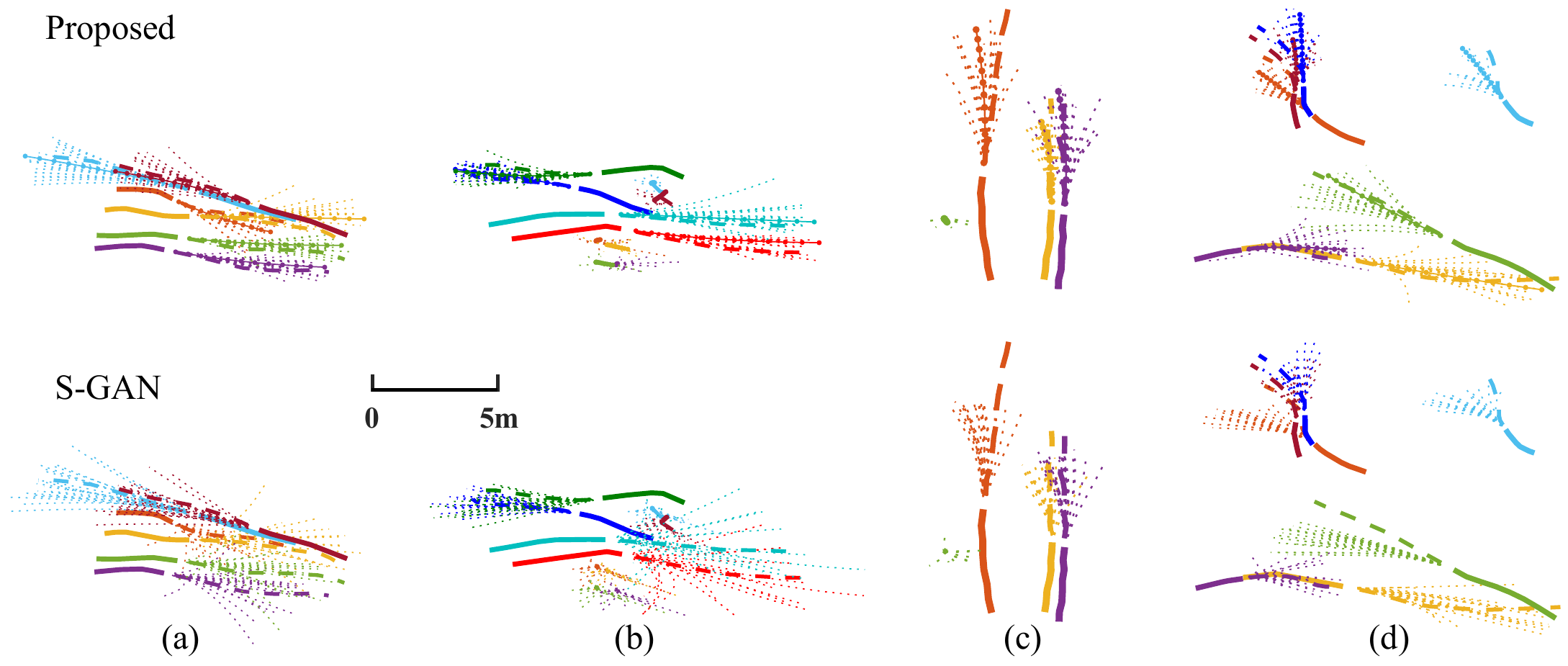}
  \caption{Examples for generated human trajectories visualization for S-GAN and our model across several scenes. The observed trajectories are shown in solid lines, ground truth future trajectories are shown in wide dashed lines, generated 20 samples per model are shown in thin dashed lines. The dot-dashed lines denote the predictions of our VAE based model by applying the mean value ($\mu_z$) of the distribution. Different humans are denoted by different colors. \label{fig:visual} }
\end{figure*}

In order to better understanding the benefits of our model in capturing social interactions between humans, we visualize several examples of the generated trajectories across testing sets as shown in Fig.\ref{fig:visual}.

 From the four examples, we can see that the predictions of our model generally have lower variance than S-GAN, which means we can generate model in a more efficient way. Also, the examples show that our model better captures the interactions of pedestrians walking in the crowds which obtain more accurate predictions (as shown in (d)). It is clear to see that our model generates more realistic predictions avoiding collisions as shown in example (b). Besides, S-GAN tends to predict slower motion in dataset HOTEL (as shown in (c)).

For qualitative results of the ablation study, please refer to Fig. 3 in supplementary file. We can observe consistent results with the quantitative evaluation. The proposed full model make more accurate and realistic predictions.

\section{Conclusion}

In this paper, we propose a novel VAE based generative model for trajectory prediction which outperforms state-of-the-art methods.
We introduce graph convolutional networks (GCNs) for efficient crowd interaction aggregation. 
Furthermore, we provided coherent motion information for the trajectory prediction datasets. 
The coherent motion labels that significantly enrich the social information for the commonly used datasets (ETH and UCY) will be released to the research community later. 
Then we incorporated the coherent motion information, which contains rich information about implicit social relationship among the humans, into our methods.
We show that the introduction of GCNs and coherent motion information significantly improve the performance for accurate trajectory prediction.

\bibliography{egbib}
\end{document}


\maketitle
\section{Coherent motion clustering}
\subsection{Parameters for coherent motion clustering}
Table \ref{tab:para} shows the parameters used for coherent filtering (CF) as well as DBSCAN. The Coherent Filtering methods are sensitive to the parameters chosen which are carefully tuned for for each dataset. As coherent filtering can easily detect coherent motions in dense environments and induce false positives, we set a larger frame window size to ensure accuracy. Besides, the angular difference is limited to a smaller value for DBSCAN for accurate detection. The typical DBSCAN applies euclidean distance as the distance function. However, we considered the angular distance, lateral distance as well as longitudinal distance for better coherent motion clustering. 

\begin{table}[!hbt]
\centering
\begin{tabular}{|l|ccc|cccc|}
\hline
\multirow{2}{*}{Dataset}                                            & \multicolumn{3}{c|}{coherent filtering} & \multicolumn{4}{c|}{DBSCAN}                     \\ \cline{2-8} 
                                                                    & $d$+2    & $K_{max}$   & $\lambda$   & $\theta$ & $s_{lateral}$ & $s_{longitudinal}$ & minPts \\ \hline
\begin{tabular}[c]{@{}l@{}}ETH, HOTEL, \\ ZARA1, ZARA2\end{tabular} & 5        & 5           & 0.8         & 0.5      & 2           & 5                & 2      \\
UNIV                                                                & 8        & 5           & 0.8         & 0.2      & 2           & 5                & 2      \\ \hline
\end{tabular}
\caption{Parameters used for coherent motion clustering. $d$+2 frames indicates the frame window size. $K_{max}$ indicates the maximum number of nearest neighbors. The coherent filtering considers $K$ nearest neighbors. We set $K = min(K_{max}, \text{ Neighborhood size})$. $\lambda$ is the threshold. $\theta$ is the angle distance and the unit is radian. $s_{lateral}$ is the lateral distance of the potential neighbors to the pedestrian considered. $s_{longitudinal}$ is the longitudinal distance. The unit is meter. minPts is the minimum points.}
\label{tab:para}
\vspace{-1.5em}
\end{table}

\begin{figure*}
  \centering
  \includegraphics[width=0.9\columnwidth]{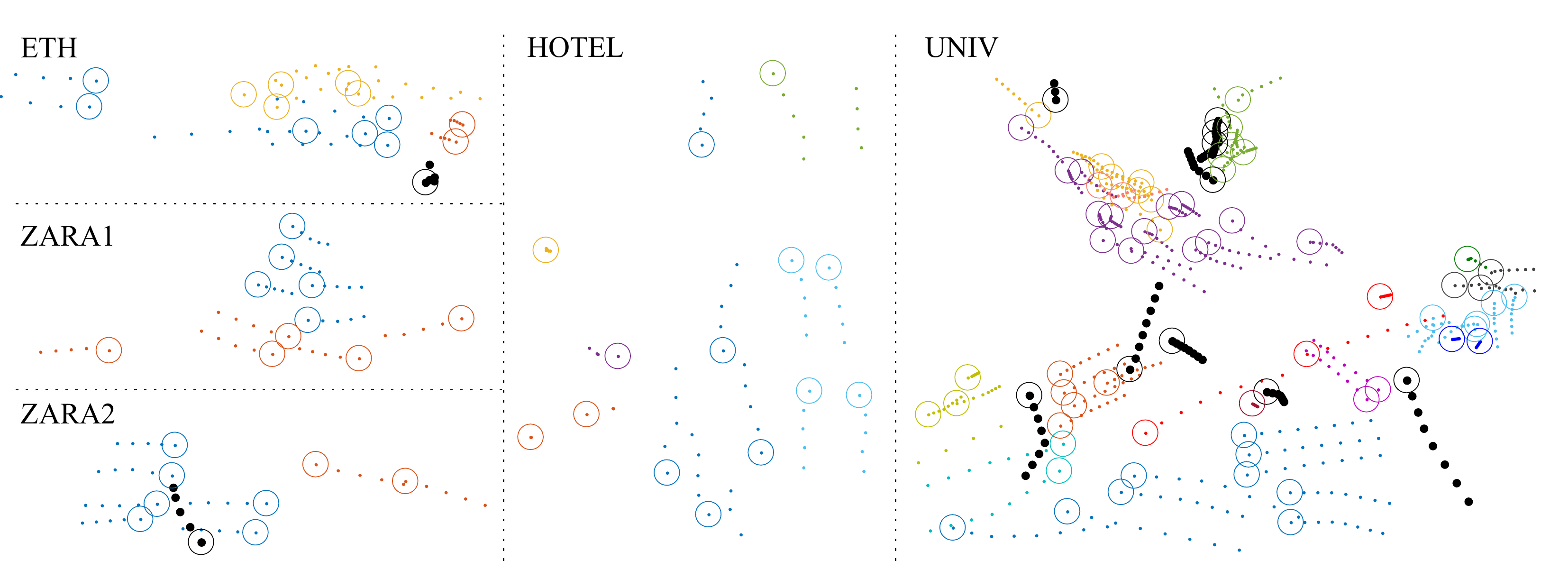}
  \caption{Representative examples and coherent motion clustering results for five datasets. Same color denotes the same group. Black color denotes individual that has no coherence with others. Circles show the current position and dots show the trajectory history used for clustering. We can see that coherent motions for both group of humans and individual humans are detected. Best viewed in color. \label{fig:coherent} }
\end{figure*}

\subsection{Coherence detection results}
Typical examples of the coherent motion detection for each dataset can be seen in Fig. \ref{fig:coherent}. We can see good performance of the coherent motion detection. 

\begin{figure}[t!]
  \centering
  \includegraphics[width=0.9\columnwidth]{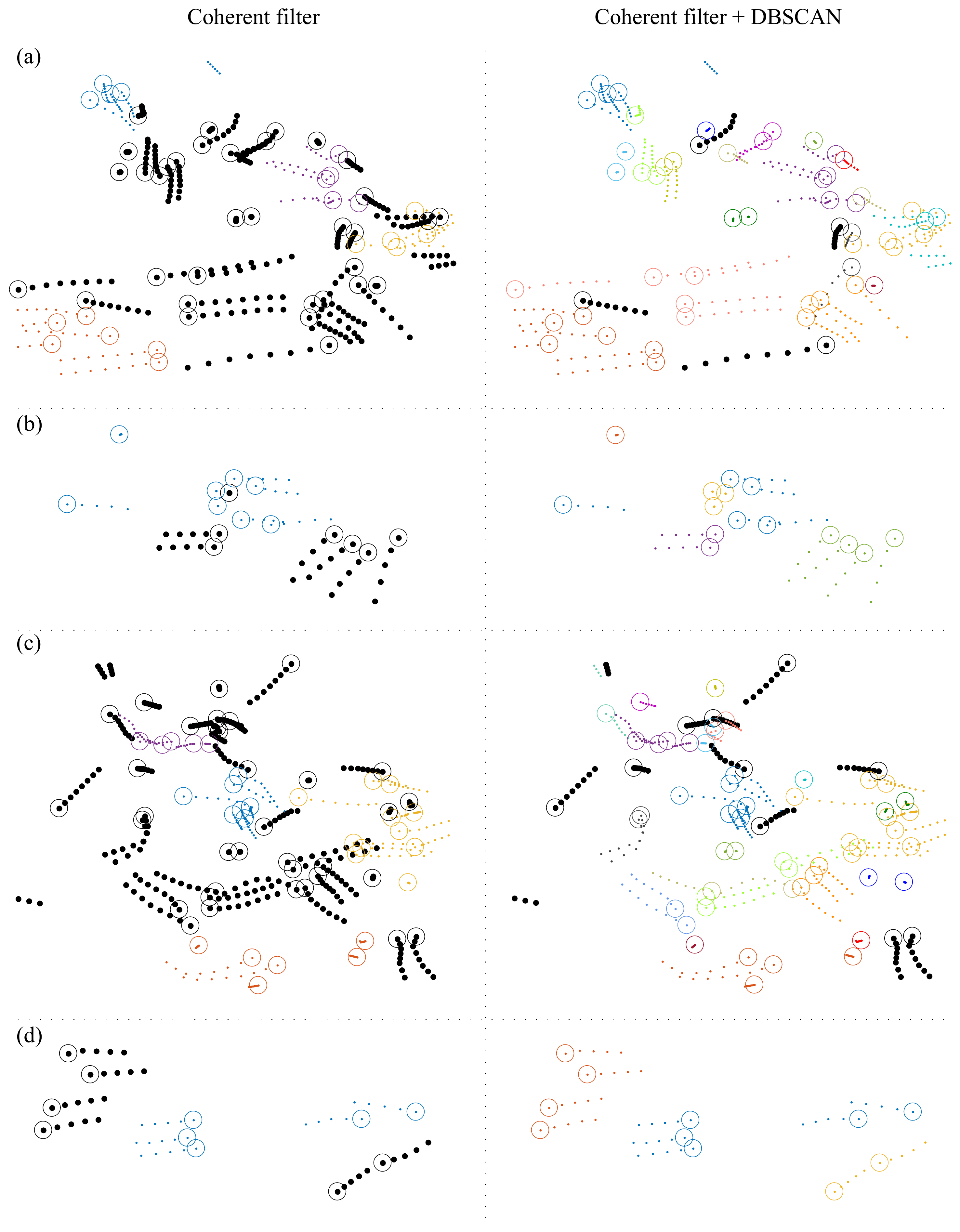}
  \caption{Representative examples for different coherent motion clustering methods. Same color denotes the same group. Black color denotes individual that has no coherence with others. Circles show the current position and dots show the trajectory history used for clustering. Best viewed in color. \label{fig:dataset} }
  \vspace{-1.5em}
\end{figure}

Fig. \ref{fig:dataset} shows the comparison of coherent motion detection between coherent filtering \cite{zhou2012coherent} and ours (leveraging DBSCAN clustering to compensate the drawback of coherent filtering).
The clustering results clearly demonstrate the performance of coherent filtering being applied to the trajectory datasets. It performed well for detecting some coherent motions in dense crowds despite the difference in the motion directions and the space separations. However, it showed poor performance for detecting coherence in sparse trajectory datasets that consist of small groups. It can be observed that pedestrians with similar moving pattern are labeled as with no coherent motion when the number of coherent pedestrians is small. This caused the low labeling rate shown in Table \ref{tab:lable_rate}, e.g. in the dataset HOTEL, of all the motions, only around 10\% are labeled as coherent motions. Besides, some static pedestrians are mis-labeled (Fig. \ref{fig:dataset}b) and become false positives for some clusters. 

To compensate, we applied the DBSCAN to detect the small pedestrian clusters. As shown in Fig. \ref{fig:dataset}, small group of pedestrians with similar moving directions are detected and labeled as the same motion cluster. Through this, the percentages of labeled coherent motions over all motions were increased to a reasonable value. 
To better show the improvement of the clustering methods, we compared the Fr\'{e}chet distance \cite{bian2018survey} of trajectory pairs of inter or intra groups clustered by coherent filtering alone or with DBSCAN. The results are shown in Table \ref{tab:similarity}. We can observe that with better coherence clustering on small groups, the Fr\'{e}chet distance of coherent trajectories classified by coherent filtering and DBSCAN becomes smaller and it becomes larger for trajectories with little coherence. A lower Fr\'{e}chet distance of two trajectories denotes higher similarity. It indicates improved coherence clustering of our proposed clustering method. 
\begin{table}[ht!]
\centering
\begin{tabular}{|l|c|c|}
\hline
Dataset & CF & CF + DBSCAN \\ \hline
ETH     & 41.0\%          & 77.3\%                   \\
HOTEL   & 12.4\%          & 77.6\%                   \\
UNIV    & 35.0\%          & 80.6\%                   \\
ZARA1   & 38.9\%          & 83.9\%                   \\
ZARA2   & 45.6\%          & 89.1\%                   \\ \hline
\end{tabular}
\caption{Percentage of labeled coherent motions over all motions. }
\label{tab:lable_rate}
\end{table}

\begin{table}[ht!]
\centering
\begin{tabular}{|l|c|c|c|c|}
\hline
                          & \multicolumn{2}{c|}{coherent filtering}                      & \multicolumn{2}{c|}{coherent filtering + DBSCAN}             \\ \cline{2-5} 
\multirow{-2}{*}{Dataset} & Intra Group                 & Inter Group                 & Intra Group                 & Inter Group                 \\ \hline
ETH                       & 3.58                        & 7.30                        & 3.21                        & 8.59                        \\
HOTEL                     & 4.10                        & 5.08                        & 2.90                        & 5.69                        \\
UNIV                      & 2.82                        & 7.28                        & 2.54                        & 7.67                        \\
ZARA1                     & 3.60                        & 5.59                        & 2.73                        & 7.64                        \\
ZARA2                     & 3.57                        & 5.37                        & 1.70                        & 6.06                        \\ \hline
AVG     &  3.53 &  6.12 & 2.62 &  7.13 \\ \hline
\end{tabular}
\caption{Similarity of intra group trajectories and inter group trajectories for the two coherent detection methods. Here we use Fr\'{e}chet distance to measure the similarity between trajectories.}
\label{tab:similarity}
\vspace{-1em}
\end{table}

\subsection{Prediction performance with different coherent detection}
Table \ref{tab:comparison} shows the prediction error (ADE and FDE) of models with different coherent motion detection methods.  Corresponding to the coherent motion detection performance,  the trajectory prediction model with coherent motions detected by CF and DBSCAN achieves better prediction performance.
\begin{table}[]
\resizebox{\columnwidth}{!}{
\begin{tabular}{|c|ccccc|c|}
\hline
                                                                   & ETH       & HOTEL     & UNIV      & ZARA1     & ZARA2     & AVG       \\ \hline
\begin{tabular}[c]{@{}c@{}}GCN+Group \\ (CF alone)\end{tabular}          & 0.71/1.28 & 0.37/0.76 & 0.55/1.19 & 0.34/0.72 & 0.32/0.68 & 0.46/0.93 \\
\begin{tabular}[c]{@{}c@{}}GCN+Group \\ (CF + DBSCAN)\end{tabular} & 0.70/1.26 & 0.37/0.75 & 0.53/1.16 & 0.34/0.71 & 0.31/0.67 & 0.45/0.91 \\ \hline
\end{tabular} }
\caption{Prediction performance comparison of models with different coherent motion clustering methods.}
\label{tab:comparison}
\vspace{-1em}
\end{table}

\section{Qualitative results of the ablation study for trajectory prediction}



\begin{figure}[t!]
  \centering
  \includegraphics[width=0.9\columnwidth]{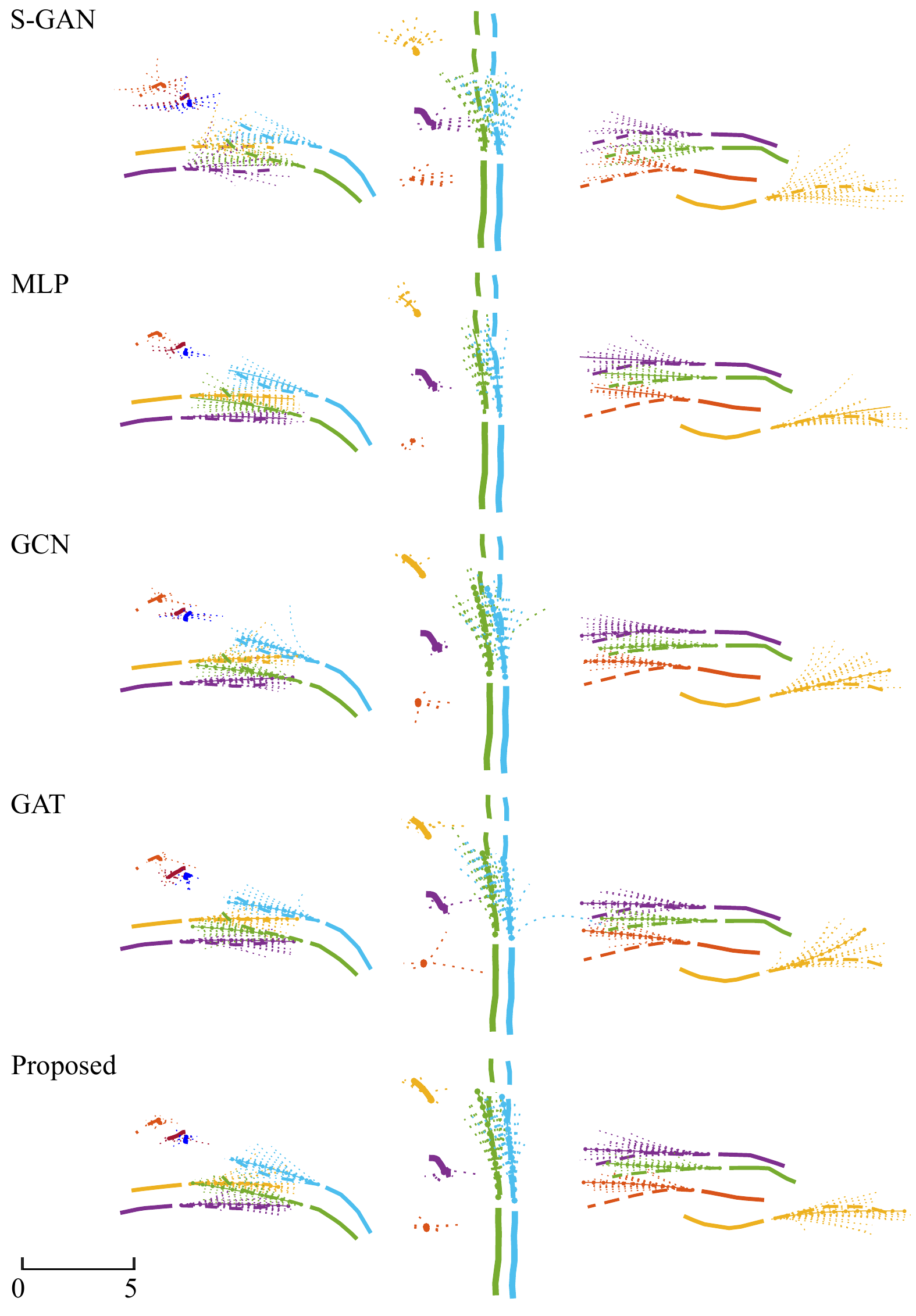}
  \caption{Examples for predicted trajectories visualization for different models. The observed trajectories are shown in solid lines, ground truth future trajectories are shown in wide dashed lines, generated 20 samples per model are shown in thin dashed lines. The dot-dashed lines denote the predictions of our VAE based model by applying the mean value ($\mu_z$) of the distribution. Different humans are denoted by different colors.  \label{fig:abla} }
\end{figure}

Figure \ref{fig:abla} shows pedestrian trajectory prediction results for different models. We can observe consistent results with the quantitative evaluation. 
When compared to S-GAN, we can see that our models often generate more accurate and efficient predictions with lower variance.
We also observed that model using MLP tested in dataset HOTEL and UNIV tends to predict slower motion of humans than the real situations, which is similar to the performance of S-GAN. Model utilizing GAT is more likely to have unexpected predictions like sharp turns shown in the second column of the figure.
Among our models, we can see the proposed full model make more accurate and realistic predictions.

\bibliography{egbib}